\documentclass[graybox]{styles/svmult}

\usepackage{mathptmx}       
\usepackage{helvet}         
\usepackage{courier}        
\usepackage{type1cm}        
\usepackage{makeidx}         
\usepackage{graphicx}        
\usepackage{multicol}        
\usepackage[bottom]{footmisc}
\usepackage{hyperref}
\usepackage{tikz,booktabs}
\usepackage{subfig}
\usepackage[utf8]{inputenc}
\usepackage[american]{babel}
\usepackage[autostyle, english = american]{csquotes}
\usepackage{siunitx}
\usepackage{lipsum}
\usepackage{bookmark}
\usepackage{placeins}
\usepackage{microtype}
\MakeOuterQuote{"}

\makeatletter
\providecommand*{\toclevel@title}{0}
\def\toclevel@author{1000}
\makeatother

\usepackage{titlesec}

\makeindex             

\begin{document}

\title*{Yeah, Right, Uh-Huh: A Deep Learning Backchannel Predictor}
\author{Robin Ruede, Markus Müller, Sebastian Stüker, Alex Waibel}
\institute{
Robin Ruede \and \email{robin.ruede@student.kit.edu} \\
Markus Müller \and \email{m.mueller@kit.edu} \\
Sebastian Stüker \and \email{sebastian.stueker@kit.edu}
\at Karlsruhe Institute of Technology, Institute for Anthropomatics and Robotics, Germany
\and
Alex Waibel \and \email{alexander.waibel@kit.edu}
\at Karlsruhe Institute of Technology, Institute for Anthropomatics and Robotics, Germany\\
Carnegie Mellon University, InterACT, Pittsburgh, PA, USA
}
%
%
%
\maketitle
\abstract{
Using supporting backchannel (BC) cues can make human-computer interaction more social.
BCs provide a feedback from the listener to the speaker indicating to the speaker that he is still listened to.
BCs can be expressed in different ways, depending on the modality of the interaction, for example as gestures or acoustic cues.
In this work, we only considered acoustic cues.
We are proposing an approach towards detecting BC opportunities based on acoustic input features like power and pitch.
While other works in the field rely on the use of a hand-written rule set or specialized features, we made use of artificial neural networks.
They are capable of deriving higher order features from input features themselves.
In our setup, we first used a fully connected feed-forward network to establish an updated baseline in comparison to our previously proposed setup.
We also extended this setup by the use of Long Short-Term Memory (LSTM) networks which have shown to outperform feed-forward based setups on various tasks.
Our best system achieved an F1-Score of 0.37 using power and pitch features. Adding linguistic information using word2vec, the score increased to 0.39.
}
\section{Introduction}
%
%
With dialog speech technology increasingly entering the mainstream of our every day lives (Siri, Cortana, Alexa, \dots), there is a
growing interest in dialog systems that are not only utilitarian (to answer questions or carry out tasks), but also to entertain and
to be social.  Humanoid robots, interactive toys, virtual assistants and even virtual psychiatrists and pets attempt to add an
emotional and social dimension to human interaction that may go beyond improving the user experience of existing dialog systems,
and thus require increasingly skillful and adept social interaction.
Social dialogs are, however, much less well understood than goal directed ones.  They do not aim for a particular outcome
other than the more indirect goals of growing a mutual understanding, empathizing, bonding and entertaining between humans.

In the present paper, we are proposing a neural network based system to generate a social response.  Our first attempt in this
regard aims to predict a suitable social response, when human speakers take “the floor” and are sharing thoughts and experiences.
The so-called “backchannel” (BC) involves short phrases (“uh-huh”, “hum", “yeah", “right”, etc.) whose role is to signal to another
speaker that one is listening and paying attention.  Further extensions also empathize, confirm, approve or disapprove.
In conversational speech, BCs complement turn taking where more rapid questions and responses are exchanged.  Despite
its simple function, however, the BC is surprisingly complex: It must be chosen properly, timed correctly and placed at appropriate
intervals.  It also responds to content, emotion and discourse state.

In this paper, we describe a neural network approach to learning the production of proper BC cues.  We will focus on
short phrasal BC cues during longer stretches of conversational speech, where another speaker has taken the floor.
Appropriate prediction of backchanneling is learned from human conversation and includes acoustic and linguistic features.
In our work, we use recurrent neural networks to learn the choice and placement of appropriate BC cues from
conversational data (Switchboard).
Special attention is given to producing “causal” backchanneling, i.e., so that the generation of a BC can be produced in
real-time systems with information of the past.

This paper is organized as follows: In the next Section, we provide an overview of related work. In Section \ref{sec:extraction}, we describe our approach in detail, followed by an overview of the experimental setup in Section \ref{experiments}. The results of the experiments are presented in Section \ref{results}. This paper concludes with an outlook to future work in Section \ref{sec:outlook}.
\section{Related Work}
\label{sec:rel_work}
Different approaches towards BC prediction have been proposed in the past.
They are based on different types of predictors and use a wide variety of input modalities.
These modalities include acoustic features like pause and pitch, but also visual cues like head movement.
In addition to these direct features, additional information sources like language models or part of speech tagging exist.

Many approaches are rule based.
\cite{eemcs18627} proposed a method that uses acoustic features.
The authors state that the most important acoustic phenomena for BC prediction occur right before a BC.
As features, they used pause information, as well as pitch (falling or rising slope).
They conducted their experiments on a Dutch corpus and report that the most important feature in their work is the duration of the pause.
\cite{ward2000prosodic} proposed a similar approach triggering BCs at low pitch and pause regions in English and Japanese. 
But building a rule based system might prove difficult as these rules have to be manually created, which is a time-consuming and difficult to generalize.
Other works included data-driven methods in which a classifier is trained and the output of this classifier is then post-processed.
\cite{Morency2010} proposes an approach that incorporates sequential probabilistic models like Hidden Markov Models or Conditional Random Fields.
They used a set of features including eye gaze and several features derived from the audio signal, e.g., downslopes in pitch or certain types of volume changes.
In another approach, predicting different types of BC was attempted \cite{kawahara2016prediction}.
Detecting BCs in real-time was also proposed \cite{schroder2012building} in the past.

There exists another category of systems that make use of artificial neural networks (ANNs).
Being a data-driven method, NNs do not require handwritten rules.
They have shown to be a versatile tool with the ability to learn relevant features automatically. 
A first approach towards detecting speech acts (including BCs) was proposed by Ries \cite{ries1999hmm}.
He used an NN in combination with an HMM.
Stolcke also proposed NN based methods for modelling dialogue acts \cite{stolcke1998dialog,stolcke2000dialogue}.
In the past, we also proposed an NN based approach \cite{mueller} that was mainly data-driven, requiring only minimal post-processing of the network outputs.
In this first approach, we used a very basic ANN based setup, which we now refined.

The objective evaluation of systems for BC prediction is difficult because BC behaviour is very speaker-dependent and subjective.
As an objective measurement, the use of the F1-Score has been established.
\cite{kok2012survey} provides a comparison of different approaches for evaluation.
In addition to objective measures, user studies are also a possibility to evaluate BC systems, like we did in the past \cite{mueller}.
A general study about the occurrence of BCs with respect to their role in facilitating attentive listening also exists \cite{kawahara2015toward}.
\section{Backchannel Prediction}
\label{sec:extraction}
\subsection{BC Utterance Selection}
\label{sec:extractio:subsec:bc-utterance-selection}
There are different kinds of phrasal BCs, they can be non-committal, positive, negative, questioning, et cetera.
To simplify the problem of predicting BCs, we only try to predict the trigger
times for any type of BC, ignoring the distinction between different kinds of responses.
%
\subsection{Feature Selection}\label{feature-selection}
A neural network is able to learn advantageous feature representations on its own.
Hence, feeding the absolute pitch and power (signal energy) values for a given time context enables the network
to automatically extract the relevant information such as pitch slopes and pause triggers, as used in related research \cite{Morency2010}.
In addition to pitch and power, we also evaluated using other acoustic features such as the fundamental frequency variation (FFV) \cite{laskowski2008fundamental} and the Mel-frequency cepstral coefficients (MFCCs).
Finally, we tried adding an encoding of the speakers' word history before the listener backchannel using word2vec \cite{mikolov_efficient_2013} to assess whether our setup benefits from multimodal input features.
\subsection{Training and Neural Network Design}\label{training}
We assumed to have two separate, but synchronized audio channels and corresponding transcripts: One for the speaker and one for the listener. We needed to decide which areas of audio to use to train the network.
As we wanted to predict the BCs in an online fashion without using future information, we needed to train the network to detect segments of audio from the speaker track that would potentially cause a
BC in the listener track.
We chose the beginning of the BC utterance as an 
anchor and used a fixed context before that as the positive prediction area.
We also needed to choose negative examples, so
the network would not be biased to always predict a BC.
We did this by selecting the range a few seconds before each BC, because in that area
the listener explicitly decided not to give a backchannel response yet. This resulted in a fully balanced training dataset.

We initially used a feed forward network architecture.
The input layer consists of all the chosen features over the previously selected fixed time context.
The output layer has two softmax neurons representing the "categories" [BC, non-BC]. We used back-propagation to train the network on the outputs [1, 0] for BC and [0, 1] for non-BC prediction areas. We only need to consider one of these outputs because the softmax function guarantees that they add up to one. We evaluated multiple different combinations of network depths and neuron counts. An example of the architecture with two hidden layers can be seen in \autoref{fig:nn}. 
\tikzstyle{layer}=[draw=black,fill=black!30]
\tikzstyle{layerlid}=[draw=black,fill=green!30]
\tikzstyle{dots}=[draw=black,fill=black]
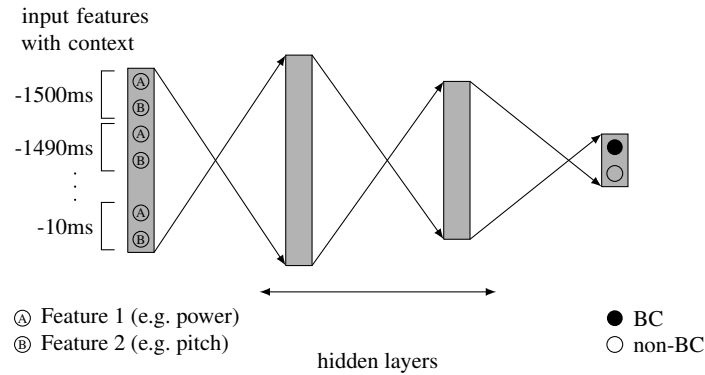
\begin{figure}[htbp]
    \tikzset{>=latex}
  \centering
  \begin{tikzpicture}[scale=0.7]
  \def \lha {2}
  \def \lhb {1.5}
  \fill[layer] (0,-1.75) -- (0.5,-1.75) coordinate(l1br) -- (0.5,1.75) coordinate(l1tr) -- (0,1.75) -- (0,-1.75);
  \fill[layer] (3,-\lha) coordinate(l2bl) -- (3.5,-\lha) coordinate(l2br) -- (3.5,\lha) coordinate(l2tr) -- (3,\lha) coordinate(l2tl) -- (3,-\lha);
  \fill[layer] (6,-\lhb) coordinate(l3bl) -- (6.5,-\lhb) coordinate(l3br) -- (6.5,\lhb) coordinate(l3tr) -- (6,\lhb) coordinate(l3tl) -- (6,-\lhb);
  \fill[layer] (9,-0.5) coordinate(o1b) -- (9.5,-0.5) -- (9.5,0.5) --
  (9,0.5) coordinate(o1t) -- (9,-0.5);
  \draw[draw=black] (0.25,1.5) circle (0.15) node {\tiny A};
  \draw[draw=black] (0.25,1) circle (0.15) node {\tiny B};
  \draw[draw=black] (0.25,0.5) circle (0.15) node {\tiny A};
  \draw[draw=black] (0.25,0) circle (0.15) node {\tiny B};
  \draw[draw=black] (0.25,-1) circle (0.15) node {\tiny A};
  \draw[draw=black] (0.25,-1.5) circle (0.15) node {\tiny B};

  \node[left,text width=2cm] at (1,2.5) {input features with context};
  \draw (-0.25,1.7) -- (-0.5,1.7) -- (-0.5,0.8) -- (-0.25,0.8) (-0.25,0.7) -- (-0.5,0.7) -- (-0.5,-0.2) -- (-0.25,-0.2) (-0.25,-0.8) -- (-0.5,-0.8) -- (-0.5,-1.7) -- (-0.25,-1.7);
  \draw (-1,-0.25) circle (0.01) ++(0,-0.25) circle (0.01) ++(0,-0.25) circle (0.01);
  \node[left] at (-0.5,1.25) {-1500ms};
  \node[left] at (-0.5,0.25) {-1490ms};
  \node[left] at (-0.5,-1.25) {-10ms};
  
  \draw[draw=black] (-2,-3) circle (0.15) node {\tiny A} +(0.2,0) node [right](power) {Feature 1 (e.g. power)};
  \draw[draw=black] (-2,-3.5) circle (0.15) node {\tiny B} +(0.2,0) node[right](pitch) {Feature 2 (e.g. pitch)};
  
  \fill (9.25,0.25) circle (0.15);
  \draw (9.25,-0.25) circle (0.15);
  \fill (9.25,-3) circle (0.15) +(0.2,0) node[right](power) {BC};
  \draw (9.25,-3.5) circle (0.15) +(0.2,0) node[right](pitch) {non-BC};
  
  \draw[->] (l1br) -- (l2tl);
  \draw[->] (l1tr) -- (l2bl);
  
  \draw[->] (l2tr) -- (l3bl);
  \draw[->] (l2br) -- (l3tl);
  \draw[->] (l3tr) -- (o1b);
  \draw[->] (l3br) -- (o1t);
  \draw[<->] (l2bl) +(-0.5,-0.5) -- ++(4,-0.5);
  \node[below](hL) at (4.75,-3.5) {hidden layers};
  \end{tikzpicture}
  \caption{Example for a neural network architecture for BC prediction.\label{fig:nn}}
\end{figure}

The placement of future BCs is dependent on the timing of previous BCs.
The probability of a BC increases with longer periods without any listener feedback. To accommodate for this, we want the network to also take its previous internal state or outputs into account. We do this by modifying the above architecture to use Long-short term memory (LSTM) layers instead of feed forward layers.
\section{Experimental Setup}
\label{experiments}
\subsection{Dataset}\label{dataset}
We used the Switchboard dataset \cite{swb}, which consists of 2,438 English
telephone conversations of five to ten minutes, 260 hours in total. Pairs of participants from across the United States were encouraged to talk about a specific topic chosen randomly from 70 possibilities. Conversation partners and topics were selected so two people would only talk once with each other, and every person would only discuss a specific topic once.

These telephone conversations are annotated with transcriptions and word alignments \cite{swbalign} with a total of 390k utterances or 3.3 million words. We split the dataset randomly into 2,000 conversations for training, 200 for validation and 238 for evaluation.
We used annotations from the Switchboard Dialog Act Corpus (SwDA) \cite{swda} to
 decide which utterances to classify as BCs. The SwDA contains
categorical annotations for the utterances of about half of the data of the
Switchboard corpus.
\subsection{Extraction}\label{extraction-1}
We chose to use the top 150 most common unique utterances marked as BCs
from the SwDA. Because the SwDA is incomplete, we had to identify
utterances as BCs just by their text. We manually included some additional utterances that were missing from the SwDA
transcriptions but present in the original transcriptions, by going
through the most common utterances and manually selecting those that
seemed relevant, such as `um-hum yeah' and `absolutely'.
The most common BCs in the data set are "yeah", "um-hum", "uh-huh" and "right", adding up to 68\% of all extracted BC phrases.

To select which utterances should be categorized as BCs and
used for training, we first filtered noise and other markers such as laughter from the
transcriptions. Some utterances such as ``uh'' can be both BCs and speech disfluencies, so we only chose those that have either silence or another
BC before them. With this method a total of 15.7\% of utterances or 2.21\% of words were labelled as BCs.

We used the Janus Recognition Toolkit \cite{janus} for parts of the feature extraction (power, pitch tracking, FFV, MFCC). Features were extracted for \SI{32}{ms} frame windows with a frame shift of \SI{10}{ms}, resulting in 100 samples per feature dimension per second. Because most of the data does not change much every 10\,ms, we also test different context strides by only extracting every n-th frame. As an example, 800\,ms of context with a stride of 2 corresponds to 40 data frames.
For word2vec, we chose to also emit one frame every 10\,ms for consistency, containing the encoding of the last non-silent word that ended before or at the  time of the frame.
\subsection{Training}\label{training-1}
We used Theano \cite{theano} with Lasagne \cite{lasagne} for rapid prototyping and testing of different parameters.\footnote{Our code for extraction, training, postprocessing and evaluation will be available at \mbox{\url{https://github.com/phiresky/backchannel-prediction}}. The repository also contains a script to reproduce all of the results of this paper.} 
To evaluate different hyperparameters, we trained multiple network configurations with various context lengths (500ms to 2000ms), context strides (1 to 4 frames), network depths (one to four hidden layers), layer sizes (15 to 125 neurons), activation functions (tanh and relu), optimization methods (SGD, Adadelta and Adam \cite{adam}), weight initialization methods (constant zero and Glorot \cite{glorot}), and layer types (feed forward and LSTM).

The LSTM networks we tested were prone to overfitting quickly. We tried two methods of regularization to overcome this. The first was Dropout training, where we randomly dropped a specific portion of neuron outputs in each layer for each training batch \cite{dropout}. We evaluated dropout layer combinations from 0 to 50\% while increasing layer sizes proportionately, but this did not improve the results. The second was adding L2-Regularization with a constant factor of 0.0001. This greatly reduced overfitting and slightly improved the results.
\subsection{Postprocessing}\label{postprocessing}
We interpret the output value of the neural networks as the probability of a BC occurring at a given time. As the output is very noisy, we first apply a low-pass filter. To ensure our prediction does not use any future information, we use a Gaussian filter which is asymmetrically cut off at some multiple $c$ of the standard deviation $\sigma$ for the side that would range into the future, and offset it so the last frame is at $\pm\SI{0}{ms}$ from the prediction target time. This means the latency of our prediction increases by $c\cdot\sigma\,ms$. If we choose $c=0$, we cut off the complete right half of the bell curve, keeping the latency at 0 at the cost of accuracy of the filter.

\begin{figure}
    \centering
    \includegraphics[width=\textwidth]{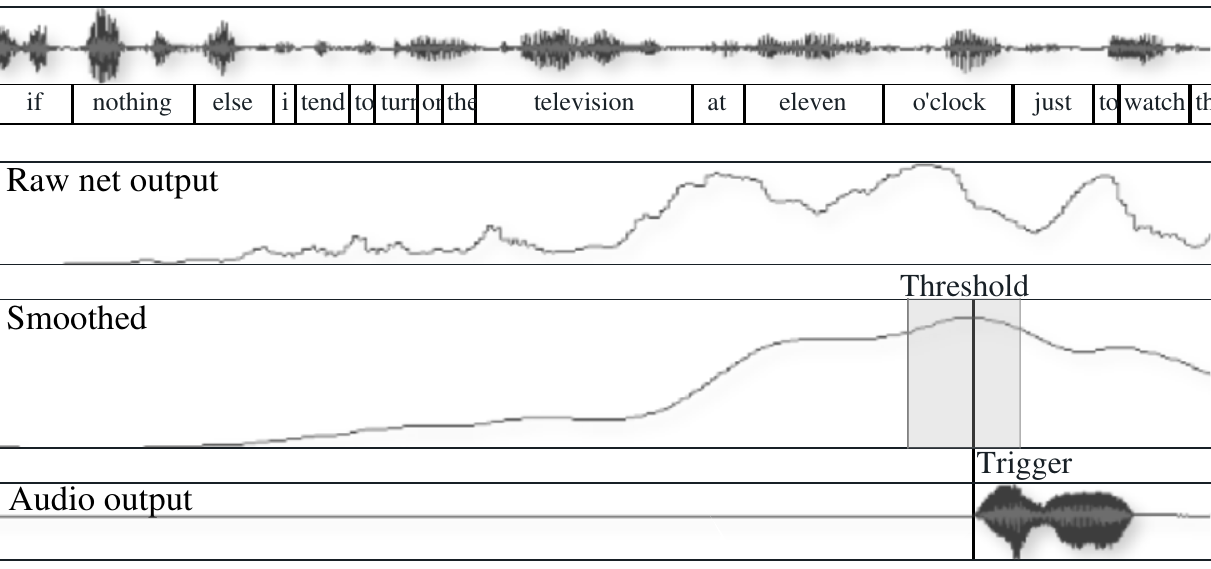}
    \caption{Postprocessing example}
    \label{fig:postproc}
\end{figure}

After the low-pass filter, we select every area for which the value exceeds a given threshold. We trigger either at the beginning of each of these areas or at their first local maximum, depending on the largest acceptable latency. This varies depending on the chosen allowed margin of error as defined in \autoref{eval-1}. 
An example of this postprocessing process can be seen in \autoref{fig:postproc}.

We determined the optimal postprocessing hyperparameters for each network configuration and allowed margin of error automatically using Bayesian optimization \cite{bayes} with the validation F1-Score as the utility function. For a margin of error of [0\,ms, +1000\,ms], the resulting standard deviation $\sigma$ ranged from \SIrange{200}{350}{ms}, and the filter cut-off ranged from $0.9\sigma\text{ to }1.4\sigma$. With this margin of error, the prediction can happen with a delay of up to one second after the ground truth. When choosing a margin of error that only allows a smaller delay such as [-200\,ms, +200\,ms], the selected standard deviation ranged from \SIrange{150}{250}{ms}, and the filter cut-off ranged from $0.0\sigma\text{ to }1.0\sigma$, causing the predicted trigger to happen earlier.
\subsection{Evaluation}\label{eval-1}
The training data contains two-sided conversations. Because the output of our predictor is only relevant for segments with just one person talking, we run our evaluation on monologuing segments.
We define a monologuing segment as the maximum possible time range in which one person is continuously \emph{talking} and the other person is continuously \emph{not talking} for at least five seconds. A person is continuously talking \emph{iff} they are only emitting utterances that are not silence or BCs. We only consider segments of a minimum length of five seconds to exclude sections of alternating conversation. This way we get segments where one participant is talking, and the other is producing backchannels without taking the turn.  

We define a prediction time as correct if it is within a given margin of error of the onset of a correct BC utterance. We did most of the testing of our predictions with an error margin of [0\,ms, +1000\,ms], but also provide results for other margins used in related research.
For comparison, we also evaluated a random predictor as a baseline. This predictor knows the correct count of BCs for a audio file and returns a uniformly distributed set of trigger times.
\section{Results}\label{results}
We use "$70 : 35$" to denote a network layer configuration of \(\text{input} \rightarrow 70\text{ neurons} \rightarrow 35\text{ neurons} \rightarrow \text{output}\).

We tested different context widths. A context width of $n\,\si{ms}$ means we use the range $[-n\,\si{ms},\allowbreak 0\,\si{ms}]$ from the beginning of the backchannel utterance. The results improved when increasing the context width from our initial value of 500\,ms. Performance peaked with a context of about 1500\,ms, as can be seen in Table \ref{varycontext}, longer contexts tended to cause the predictor to trigger too late.
We tested using only every n-th frame of input data. Even though we initially did this for performance reasons, we noticed that training on every single frame has worse performance than skipping every second frame due to overfitting. Taking every fourth frame seems to miss too much information, so performance peaks at a context stride of 2, as seen in Table \ref{varystrides}.

We tested different combinations of features, with using solely power in a first approach.
But adding prosodic features gives great improvements. Using FFV as the only prosodic feature performs worse than FFV together with the absolute pitch value. Adding MFCCs does not seem to improve performance in a meaningful way, when also using pitch, see Table \ref{varyfeatures} for more details. Note that using \emph{only} word2vec performs reasonably well, because with our method it indirectly encodes the time since the last utterance, similar to the power feature.
Table \ref{varylstm} shows a comparison between feed forward and LSTM networks. The parameter count is the number of connection weights the network learns during training. Note that LSTMs have higher performance, even with similar parameter counts.
We compared different layer sizes for our LSTM networks, as shown in Table \ref{varylayers}. A network depth of two hidden layers worked best, but the results are adequate with a single hidden layer or three hidden layers.

In \autoref{fig:final}, our final results are given for the completely independent evaluation data set. We compared the results from \cite{mueller} with our system. \cite{mueller} used the same dataset, but focused on offline predictions, meaning their network had future information available, and they evaluated their performance on the whole corpus including segments with silence and with alternating conversation. We adjusted our baseline and evaluation system to match their setup by removing the monologuing constraint described in \autoref{eval-1} and changing the margin of error to [-200\,ms, +200\,ms]. As can be seen in Table \ref{fig:mueller}, our predictor performs better.
All other related research used different languages, datasets or evaluation methods, rendering a direct comparison difficult because of slightly different tasks.

Table \ref{fig:ourbest} shows the results with our presented evaluation method. We provide scores for different margins of error used in other research. Subjectively, missing a BC trigger may be more acceptable than a false positive, so we also provide a result with balanced precision and recall. Note that a later margin center with the same margin width has higher performance because it allows more latency in the predictions, which means we can choose better postprocessing parameters as described in \autoref{postprocessing}.

\section{Conclusion and Future Work}
\label{sec:outlook}
We have presented a new approach to predict BCs using neural networks.
With refined methods for network training as well as different network architectures, we could improve the F1-Score in contrast to our previous experiments.
In addition to evaluating different hyperparameter configurations, we also experimented with LSTM networks, which lead to improved results.
Our best system achieved an F1-Score of 0.388.

We used linguistic features via word2vec only in a very basic way, assuming the availability of an instant speech recognizer by using the reference transcripts.
As low-latency speech recognition is possible \cite{niehues2016dynamic}, one of the next steps would be to combine both systems.
Further work is needed to evaluate other methods for adding word vectors to the input features and to analyze problems with our approach.
We only tested feed forward neural networks and LSTMs, other architectures like time-delay neural networks \cite{waibel1989phoneme}, also called convolutional neural networks, may also give interesting results.
Our approach of choosing the training areas may not be optimal because the delay between the last utterance of the speaker and the backchannel can vary significantly. One possible solution would be to align the training area by the last speaker utterance instead of the backchannel start. 

Because backchannels are a largely subjective phenomenon, a user study would be helpful to subjectively evaluate the performance of our predictor and to compare it with human performance in our chosen evaluation method. Another method would be to find consensus for backchannel triggers by combining the predictions of multiple human subjects for a single speaker channel as described by Huang et al. (2010) as "parasocial consensus sampling" \cite{huang2010learning}.

\begin{table}
\caption{Results on the Validation Set. All results use the following setup if not otherwise stated: LSTM, configuration: $(70 : 35)$; input features: power, pitch, ffv; context width: 1500\,ms; context frame stride: 2; margin of error: 0\,ms to +1000\,ms. Precision, recall, and F1-Score are given for the validation data set.}\label{fig:survey}
\centering
\subfloat[Results with various context lengths. Performance peaks at 1500\,ms.]{
    \begin{tabular}{cccc}
    \hline\noalign{\smallskip}
    Context & Precision & Recall & F1-Score \\
    \noalign{\smallskip}\svhline\noalign{\smallskip}
    500\,ms & 0.219 & 0.466 & 0.298 \\
    1000\,ms & 0.280 & 0.497 & 0.358 \\
    1500\,ms & 0.305 & 0.488 & \bf{0.375} \\
    2000\,ms & 0.275 & 0.577 & 0.373 \\
    \noalign{\smallskip}\hline\noalign{\smallskip}
    \end{tabular}
    \label{varycontext}
}
\qquad
\subfloat[Results with various context frame strides.]{
    \begin{tabular}{cccc}
    \hline\noalign{\smallskip}
    Stride & Precision & Recall & F1-Score \\
    \noalign{\smallskip}\svhline\noalign{\smallskip}
    1 & 0.290 & 0.490 & 0.364 \\
    2 & 0.305 & 0.488 & \bf{0.375} \\
    4 & 0.285 & 0.498 & 0.363 \\
    \noalign{\smallskip}\hline\noalign{\smallskip}
    \end{tabular}
    \label{varystrides}
}

\centering
\subfloat[Results with various input features, separated into only acoustic features and acoustic plus linguistic features.]{
    \begin{tabular}{lccc}
    \hline\noalign{\smallskip}
    Features & Precision & Recall & F1-Score \\
    \noalign{\smallskip}\svhline\noalign{\smallskip}
    power & 0.244 & 0.516 & 0.331 \\
    power, pitch & 0.307 & 0.435 & 0.360 \\
    power, pitch, mfcc & 0.278 & 0.514 & 0.360 \\
    power, ffv & 0.259 & 0.513 & 0.344 \\
    power, ffv, mfcc & 0.279 & 0.515 & 0.362 \\
    power, pitch, ffv & 0.305 & 0.488 & \bf{0.375} \\
    \noalign{\smallskip}\hline\noalign{\smallskip}
    word2vec$_{dim=30}$ & 0.244 & 0.478 & 0.323 \\
    power, pitch, word2vec$_{dim=30}$ & 0.318 & 0.486 & 0.385 \\
    power, pitch, ffv, word2vec$_{dim=15}$ & 0.321 & 0.475 & 0.383 \\
    power, pitch, ffv, word2vec$_{dim=30}$ & 0.322 & 0.497 & \bf{0.390} \\
    power, pitch, ffv, word2vec$_{dim=50}$ & 0.304 & 0.527 & 0.385 \\
    \noalign{\smallskip}\hline\noalign{\smallskip}
    \end{tabular}
    \label{varyfeatures}
}

\subfloat[Feed forward vs LSTM. LSTM outperforms feed forward architectures.]{
    \begin{tabular}{ccccc}
    \hline\noalign{\smallskip}
    Layers & Parameter count & Precision & Recall & F1-Score \\
    \noalign{\smallskip}\svhline\noalign{\smallskip}
    FF ($56 : 28$) & 40k & 0.230 & 0.549 & 0.325 \\
    FF ($70 : 35$) & 50k & 0.251 & 0.468 & 0.327 \\
    FF ($100 : 50$) & 72k & 0.242 & 0.490 & 0.324 \\
    LSTM ($70 : 35$) & 38k & 0.305 & 0.488 & \bf{0.375} \\
    \noalign{\smallskip}\hline\noalign{\smallskip}
    \end{tabular}
    \label{varylstm}
}

\subfloat[Comparison of different network configurations. Two LSTM layers give the best results.]{
    \begin{tabular}{cccc}
    \hline\noalign{\smallskip}
    Layer sizes & Precision & Recall & F1-Score \\
    \noalign{\smallskip}\svhline\noalign{\smallskip}
    $100$ & 0.280 & 0.542 & 0.369 \\
    $50 : 20$ & 0.291 & 0.506 & 0.370 \\
    $70 : 35$ & 0.305 & 0.488 & \bf{0.375} \\
    $100 : 50$ & 0.303 & 0.473 & 0.369 \\
    $70 : 50 : 35$ & 0.278 & 0.541 & 0.367 \\
    \noalign{\smallskip}\hline\noalign{\smallskip}
    \end{tabular}
    \label{varylayers}
}

\end{table}

\newcommand{\csubfloat}[2][]{%
  \makebox[0pt]{\subfloat[#1]{#2}}%
}
\captionsetup[subfigure]{width=\textwidth}
\begin{table}
\centering
\caption{Final best results on the evaluation set (chosen by validation set)}\label{fig:final}

\csubfloat[Comparison with previous research. \cite{mueller} did their evaluation without the constraints defined in \autoref{eval-1}, so we adjusted our baseline and evaluation to match their setup.]{
    \begin{tabular}{lccc}
    \hline\noalign{\smallskip}
        Predictor & Precision & Recall & F1-Score \\
        \noalign{\smallskip}\svhline\noalign{\smallskip}
        Baseline (random) & 0.042 & 0.042 & 0.042 \\
        Müller et al. (offline) \cite{mueller} & -- & -- & 0.109 \\
        Our results (offline, context of \SIrange{-750}{750}{ms}) & 0.114 & 0.300 & \bf{0.165} \\
        Our results (online, context of \SIrange{-1500}{0}{ms}) & 0.100 & 0.318 & 0.153 \\
    \noalign{\smallskip}\hline\noalign{\smallskip}
    \end{tabular}
    \label{fig:mueller}
}

\csubfloat[Results with our evaluation method with various margins of error used in other research \cite{survey}. Performance improves with a wider margin width and with a later margin center.]{
    \begin{tabular}{clccc}
    \hline\noalign{\smallskip}
        Margin of Error & Constraint & Precision & Recall & F1-Score \\
        \noalign{\smallskip}\svhline\noalign{\smallskip}
        \SIrange{-200}{+200}{ms} && 0.172 & 0.377 & 0.237 \\
        \SIrange{-100}{+500}{ms} &&	0.239 & 0.406 & 0.301 \\
        \SIrange{-500}{+500}{ms} && 0.247 & 0.536 & 0.339 \\
    \hline\noalign{\smallskip}
        \SIrange{0}{+1000}{ms} & Baseline (random, correct BC count) & 0.111 & 0.052 & 0.071 \\
         & Baseline (random, 8x correct BC count) & 0.079 & 0.323 & 0.127 \\
         & Balanced Precision and Recall & 0.342 & 0.339 & 0.341 \\
         & Best F1-Score (only acoustic features) & 0.294 & 0.488 & 0.367 \\
         & Best F1-Score (acoustic and linguistic features) & 0.312 & 0.511 & \bf{0.388} \\
    \noalign{\smallskip}\hline\noalign{\smallskip}
    \end{tabular}
    \label{fig:ourbest}
}

\end{table}

\begin{acknowledgement}
This work has been conducted in the SecondHands project
which has received funding from the European Union’s
Horizon 2020 Research and Innovation programme (call:H2020-
ICT-2014-1, RIA) under grant agreement No 643950.
\end{acknowledgement}

%
%
%


\FloatBarrier

\bibliographystyle{styles/spmpsci}
\bibliography{bib}
\end{document}